\documentclass{article}
\usepackage{ijcai22}

\usepackage{times}
\usepackage{soul}
\usepackage{url}
\usepackage[hidelinks]{hyperref}
\usepackage[utf8]{inputenc}
\usepackage[small]{caption}
\usepackage{graphicx}
\usepackage{amsmath}
\usepackage{amsthm}
\usepackage{booktabs}
\usepackage{algorithm}
\usepackage{algorithmic}
\usepackage{subfigure}
\usepackage{multirow}
\title{A Higher-Order Semantic Dependency Parser}

\iftrue
\author{
Bin Li$^{1,2}$\and
Yunlong Fan$^{1,2}$ \and
Yikemaiti Sataer$^{1,2}$\and
Zhiqiang Gao$^{1,2}$\footnote{Contact Author}
\affiliations
$^1$School of Computer Science and Engineering, Southeast University, Nanjing 210096, China\\
$^2$Key Laboratory of Computer Network and Information Integration (Southeast University), Ministry of
Education, China
\emails
\{lib, fanyunlong, yikmat, zqgao\}@seu.edu.cn
}
\fi

\begin{document}

\maketitle

\begin{abstract}
Higher-order features bring significant accuracy gains in semantic dependency parsing. However, modeling higher-order features with exact inference is NP-hard. 
Graph neural networks (GNNs) have been demonstrated to be an effective tool for solving NP-hard problems with approximate inference in many graph learning tasks.
Inspired by the success of GNNs, we investigate building a higher-order semantic dependency parser by applying GNNs. Instead of explicitly extracting higher-order features from intermediate parsing graphs, GNNs aggregate higher-order information concisely by stacking multiple GNN layers.
Experimental results show that our model outperforms the previous state-of-the-art parser on the SemEval 2015 Task 18 English datasets.
\end{abstract}

\section{Introduction}
Semantic dependency parsing (SDP) attempts to identify semantic relationships between words in a sentence by representing the sentence as a labeled directed acyclic graph (DAG), also known as the semantic dependency graph (SDG).
In an SDG, not only semantic predicates can have multiple
or zero arguments, but also words from the sentence
can be attached as arguments to more than
one head word (predicate), or they can be outside
the SDG (being neither a predicate nor an argument).
SDP originates from syntactic dependency parsing which aims to represent the syntactic structure of a sentence by means of a labeled tree. 

\begin{figure}[!ht]
	\centering  %图片全局居中
	\vspace{-0.35cm} %设置与上面正文的距离
	\subfigtopskip=2pt %设置子图与上面正文或别的内容的距离
	\subfigbottomskip=2pt %设置第二行子图与第一行子图的距离，即下面的头与上面的脚的距离
	\subfigcapskip=-2pt %设置子图与子标题之间的距离
	\subfigure[PSD formalism of the example sentence]{
		\label{gold annotation}
		\includegraphics[width=0.46\textwidth]{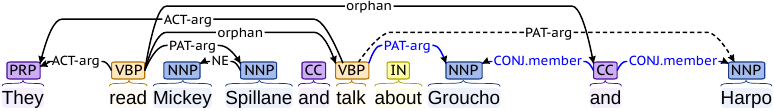}}
	\subfigure[Part of higher-order features]{
		\label{higher-order features}
		\includegraphics[width=0.46\textwidth]{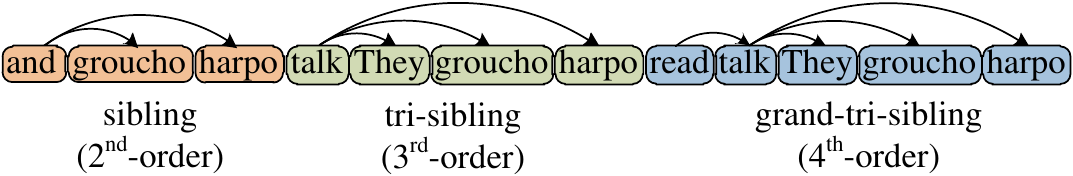}}
	\caption{Part of $2^{nd}$-order, $3^{rd}$-order, $4^{th}$-order features appeared in the example sentence.}
	\label{fig:higher_order_eg}
\end{figure}

In syntactic and semantic dependency parsing, higher-order parser generally outperforms first-order parser \cite{ji2019graph,wang2019second,zhang2020efficient}. 
The basic first-order parser scores dependency edges independently, and the higher-order parser takes relationships between the head and modifier tokens and the $K$-order neighborhoods of them into consideration. 
Here is a straightforward example (as in Figure \ref{fig:higher_order_eg}). For the sentence \emph{They read Mickey Spillane and talk about Groucho and Harpo}, its semantic dependency representation is shown in Figure \ref{gold annotation}, part of higher-order features appeared in this sentence are shown in Figure \ref{higher-order features}.
When considering the semantic dependency relationship between \emph{talk} and \emph{Harpo} (dotted edge), if we have known (1) \emph{Groucho} and \emph{Harpo} are included in the $2^{nd}$-order feature (\emph{sibling}) and (2) there is a dependency edge labeled \emph{PAT-arg} between \emph{talk} and \emph{Groucho} (blue edges), it is obvious to see that there is also a dependency edge labeled \emph{PAT-arg} between \emph{talk} and \emph{Harpo}.

Several semantic dependency parsers are presented in recent years.
Most of them are first-order parsers
\cite{du2015peking,peng2017deep,dozat2018simpler,wang2018neural,fernandez2020transition}, the rest are second-order parsers \cite{martins2014priberam,cao2017quasi,wang2019second}, higher-order parsing remains under-explored yet. The reason for this issue is that modeling higher-order features with exact inference is NP-hard. 
Higher-order features have been widely exploited in syntactic dependency parsing \cite{carreras2007experiments,koo2010efficient,ma2012fourth,ji2019graph,zhang2020efficient}.  
However, most of the previous algorithms for higher-order syntactic dependency tree parsing are not applicable to semantic dependency graph parsing.

Graph neural networks (GNNs) have been demonstrated to be an effective tool for solving NP-hard problems with approximate inference in many graph learning tasks \cite{prates2019learning,zhang2019deep,zhao2021distributed}.
GNNs aggregate higher-order information in a similar incremental manner: one GNN layer encodes information about immediate neighbors and $K$ layers encode $K$-order neighborhoods (i.e., information about nodes at most $K$ hops aways).

Inspired by the success of GNNs, we investigate building a higher-order semantic dependency parser (HOSDP) by applying GNNs in this paper. Instead of explicitly extracting higher-order features from intermediate parsing graphs, we aggregate higher-order information and bring global evidence into decoders’ final decision by stacking multiple GNN layers. We extend the biaffine parser \cite{dozat2018simpler} and employ it as the vanilla parser to produce an initial adjacency matrix (close to gold) since there is no graph structure available during testing. Two GNNs variants, Graph Convolutional Network (GCN) \cite{kipf2016semi} and Graph Attention Network (GAT) \cite{velivckovic2017graph} have been investigated in HOSDP.
Our model has been evaluated on SemEval 2015 Task 18 English Dataset which contains three semantic dependency formalisms (DM, PAS, and PSD). 
Experimental results show that HOSDP outperforms the previous best one. In addition, HOSDP shows more advantage over the baseline in the longer sentence and PSD formalism (appearing linguistically most fine-grained). 
Our code is publicly available at \url{https://github.com/LiBinNLP/HOSDP}. 

\section{Related work}
In this section, the studies of higher-order syntactic dependency parsing, semantic dependency parsing and GNNs will be summarized as follows.  
\subsection{Higher-order Syntactic Dependency Parsing}
Higher-order parsing has received a lot of attention in the syntactic dependency parsing.
% \citet{mcdonald2006online} presented a second-order parser which extends the maximum spanning tree parsing framework to incorporate second-order sibling parts.
\citet{carreras2007experiments} presented a second parser which incorporate grand-parental relationships in the dependency structure.
\citet{koo2010efficient} developed a third-order parser. They introduced grand-sibling and tri-sibling parts. 
\citet{ma2012fourth} developed a forth-order parser which utilized  grand-tri-sibling parts for fourth-order dependency parsing.
\citet{ji2019graph} used graph neural networks to captures higher-order information for syntactic dependency parsing, which was closely related to our work.
\citet{zhang2020efficient} presented a second-order TreeCRF extension to the biaffine parser. 

Higher-order features have been widely exploited in syntactic dependency parsing. However, most of the previous algorithms for higher-order syntactic dependency tree parsing are not applicable to semantic dependency graph parsing.

\subsection{Semantic Dependency Parsing}
Several SDP models are presented in the recent years. Their parsing mechanisms are either transition-based or graph-based. Most of them are first-order parsers.
\citet{wang2018neural} presented a neural transition-based parser, using a variant of list-based arc-eager transition algorithm for dependency graph parsing. 
\citet{lindemann2020fast} developed a transition-based parser for \emph{Apply-Modify} dependency parsing. They extended the stack-pointer model to predict transitions. 
\citet{fernandez2020transition} developed a transition-based parser, using Pointer Network to choose a transition between \emph{Attach-p} and \emph{Shift}.
More recently, there has been a predominance of purely graph-based SDP models.
\citet{dozat2018simpler} extended the LSTM-based syntactic parser of
\citet{dozat2017stanford} to train on and generate SDG. 
\citet{kurita2019multi} developed a reinforcement learning-based approach that iteratively applies the syntactic dependency parser to build a DAG structure sequentially. 
\citet{jia2020semi} presented a semi-supervised model based on Conditional Random Field Autoencoder to learn a dependency graph parser. 
\citet{he2020establishing} significantly improved the performance by introducing contextual string embeddings (called Flair). 

Higher-order SDP receives less attention.
\citet{martins2014priberam} developed a second-order parser which employ a feature-rich linear model to incorporate first and second-order features (arcs, siblings, grandparents and co-parents). $AD^3$ algorithm was employed for approximate decoding.
\citet{cao2017quasi} presented a quasi-second-order parser and used a dynamic programming algorithm (called Maximum Subgraph) for exact decoding. 
\citet{wang2019second} extended the parser \cite{dozat2018simpler} and managed to add second-order information for score computation and then apply either mean-field variational inference or loopy belief propagation for approximate decoding.

Higher-order parsing remains under-explored yet. The reason for this issue is that modeling higher-order features with exact inference is NP-hard. 

\subsection{Graph Neural Networks}
Recent years have witnessed great success from GNNs in graph learning.
Two main families of GNNs have been proposed, i.e., spectral methods and spatial methods. 
GCNs \cite{kipf2016semi} is a spectral-based method, which learns node representation based on graph spectral theory. 
GAT \cite{velivckovic2017graph} is a spatial-based method, which introduces the multi-head attention mechanism to learn different attention scores for neighbors when aggregating information. 

GNNs have been demonstrated to be an effective tool for solving NP-hard problems with approximate inference in many graph learning tasks.
\citet{prates2019learning} utilized GNNs to solve the Traveling Salesperson Problem, a highly relevant NP-Complete problem.
\citet{zhang2019deep} used GNNs to solve NP-hard assignment problems in image feature matching task.
\citet{zhao2021distributed} used GCNs to solve maximum weighted independent set problem, which is NP-hard. Inspired by the success of GNNs, we investigate building a higher-order semantic dependency parser by applying GNNs.

\begin{figure*}[!ht]
        \centering
        \includegraphics[width=0.8\textwidth]{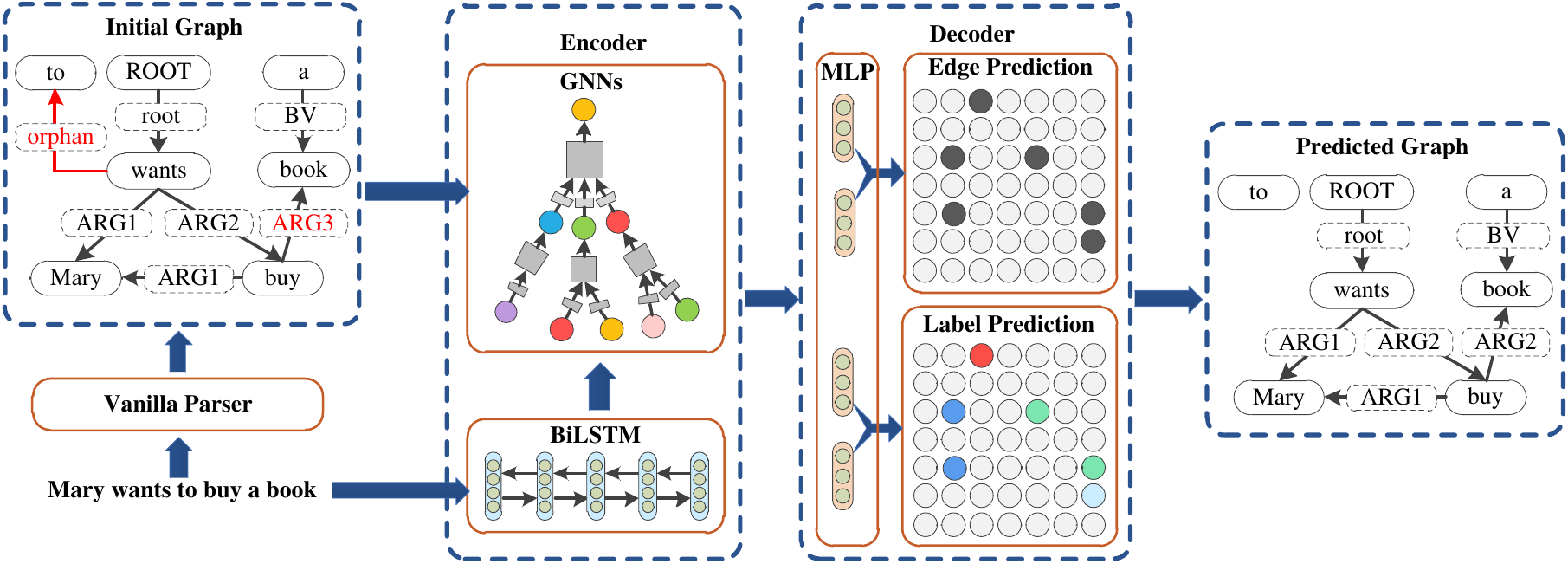}
        \caption{Overall architecture of the proposed HOSDP.}
        \label{fig:our_model}
\end{figure*}
\section{HOSDP}

HOSDP is a parser extends \citet{dozat2018simpler}. An overview of HOSDP is shown as Figure \ref{fig:our_model}. Given sentence $s$ with $n$ words $[w_1, w_2,\dots,w_n]$,  there are three steps to parse it as an SDG. Firstly, the sentence will be parsed with a vanilla SDP parser, 
producing an initial SDG. Secondly, the contextualized word representations output by long short-term memory (BiLSTM) and adjacency matrix obtained from the initial SDG will be fed into the GNN encoder, to obtain node representations which aggregate higher-order information. Finally, MLP will be used to get the hidden state representation, and then decoded by biaffine classifier to predict edge and label.

\subsection{Vanilla Parser}
We use biaffine parser \cite{dozat2018simpler} as the vanilla parser. The sentence $s$ will be parsed by the vanilla parser to obtain initial adjacency matrix.
\begin{equation}
(\Tilde{A}, \Tilde{G}) = VanillaParser(W, F)
\end{equation}
where $\Tilde{A}$ is the initial adjacency matrix; $\Tilde{G}$ is the initial SDG; $W$ denotes $n$ words and $F$ denotes features of words.

\subsection{Encoder}
We concatenate word and feature embeddings, and feed them into a BiLSTM to obtain contextualized representations.

\begin{equation}
x_i=e_i^{(word)}\oplus e_i^{(feat)}
\end{equation}

\begin{equation}
O=BiLSTM(X)
\end{equation}
where $x_i$ is the concatenation ($\oplus$) of the word and
feature embeddings of word $w_i$, $X$ represents $[x_1,x_2,\dots,x_n]$. $O=[o_1,o_2,\dots,o_n]$ is the contextualized representations of sequence $X$. 

Then we employ $K$-layer GNNs to capture higher-order information by aggregating representation of $K$-order neighborhoods. Node embedding matrix $R^{(k)}$ in $k^{th}$-layer is computed as Equation \ref{GNNLayer}:

\begin{equation}
R^{(k)}=\emph{GNNLayer}^{(k-1)}(R^{(k-1)}, \Tilde{A}) \label{GNNLayer}
\end{equation}

When \emph{GNNLayer} is implemented in \emph{GCN}, the representation of node $i$ in $k^{th}$ layer $r_i^{(k)}$ is computed as Equation \ref{GCNLayer}

\begin{equation}
r_i^{(k)}=\sigma \left(W\sum_{j\in N(i)} r_j^{(k-1)}+B r_i^{(k-1)}\right)\label{GCNLayer}
\end{equation}
where $W$ and $B$ are parameter matrices; $N(i)$ are neighbors of node $i$; $\sigma$ is active function (we use ReLU); $r_i^{(0)} = o_i$.

When \emph{GNNLayer} is implemented in \emph{GAT}, $r_i^{(k)}$ is computed as Equation \ref{GATLayer}:
\begin{equation}
r_i^{(k)}=\sigma \left(W\sum_{j\in N(i)} a_{ij}^{(m)(k-1)}r_j^{(k-1)}+B r_i^{(k-1)}\right)\label{GATLayer}
\end{equation}
where $a_{ij}^{(m)(k-1)}$ is attention coefficient of node $i$ to its neighbour
$j$ in attention head $m$ at $(k-1)^{th}$ layer.

Although higher-order information is important, GNNs would suffer from the over-smoothing problem when the number of layer is too many. So we stack 3 layers with the past experience.

\subsection{Decoder}
Decoder has two modules: edge existence prediction module and edge label prediction module. For each of the two modules, we use \emph{MLP}
to split the final node representation $R=[r_1,r_2,\dots, r_n]$ into two parts—a head representation, and a dependent representation, as shown in Equation \ref{edge-head} to \ref{label-dep}:
\begin{equation}
h_i^{(edge-head)}=MLP^{(edge-head)}(ri)\label{edge-head}
\end{equation}

\begin{equation}
h_i^{(label-head)}=MLP^{(label-head)}(ri)\label{label-head}
\end{equation}

\begin{equation}
h_i^{(edge-dep)}=MLP^{(edge-dep)}(ri)\label{edge-dep}
\end{equation}

\begin{equation}
h_i^{(label-dep)}=MLP^{(label-dep)}(ri)\label{label-dep}
\end{equation}
 
We can then use biaffine classifiers (as Equation \ref{biaff}), which are 
generalizations of linear classifiers
to include multiplicative interactions between two
vectors—to predict edges and labels, as Equation \ref{edge} and \ref{label} :

\begin{equation}
Biaff(x_1, x_2)=x_1^TUx_2 + W(x_1\oplus x_2)+b \label{biaff}
\end{equation}

\begin{equation}
s_{i,j}^{(edge)}=Biaff^{(edge)}(h_i^{(edge-dep)}, h_j^{(edge-head)})\label{edge}
\end{equation}

\begin{equation}
s_{i,j}^{(label)}=Biaff^{(label)}(h_i^{(label-dep)}, h_j^{(label-head)})\label{label}
\end{equation}
where $s_{i,j}^{(edge)}$ and $s_{i,j}^{(label)}$ are scores of 
edge existence and edge label between the word $w_i$ and $w_j$. 
$U$, $W$ and $b$ are learned parameters
of biffine classifier. For edge existence prediction module,
$U$ will be $(d*1*d)$-dimensional,  so that $s_{i,j}^{(edge)}$ 
will be a scalar. For edge label prediction, if the parser is unlabeled, 
$U$ will be $(d*1*d)$-dimensional, so that $s_{i,j}^{(label)}$ will be a scalar.
If the parser is labeled, $U$ will be $(d*c*d)$-dimensional, 
where $c$ is the number of labels, so that $s_{i,j}^{(label)}$ is a vector
that represents the probability distributions of each label.

The unlabeled parser scores each edge between
pairs of words in the sentence—these scores
can be decoded into a graph by keeping only edges
that received a positive score. The labeled parser scores
every label for each pair of words, so we simply
assign each predicted edge its highest-scoring label
and discard the rest.

\begin{equation}
\hat{y}_{i,j}^{(edge)} = \{s_{i,j}^{(edge)}>0\}
\end{equation}

\begin{equation}
\hat{y}_{i,j}^{(label)} = \arg\max s_{i,j}^{(label)} 
\end{equation}

\subsection{Learning}
We can train the system by summing the losses from the two modules, 
back propagating error to the parser.
Cross entropy function is used as as the loss function, which is computed as Equation \ref{crossentropy}:
\begin{equation}
CE(p,q)=-\sum_{x}p(x)\log q(x) \label{crossentropy}
\end{equation}

We define the loss function of edge existence prediction module 
and edge label prediction module:
\begin{equation}
\mathcal{L}^{(edge)}(\theta_2)=CE(\hat{y}_{i,j}^{(edge)}, y_{i,j}^{(edge)})
\end{equation}

\begin{equation}
\mathcal{L}^{(label)}(\theta_1)=CE(\hat{y}_{i,j}^{(label)}, y_{i,j}^{(label)})
\end{equation}
where $\theta_1$ and  $\theta_2$ are the parameters of two modules.

Then the Adaptive Moment Estimation (Adam) method is used to optimize
the summed loss function $\mathcal{L}$:
\begin{equation}
\mathcal{L}=\lambda \mathcal{L}^{(edge)} + (1-\lambda) \mathcal{L}^{(label)}
\end{equation}
where $\lambda$ is a tunable interpolation constant $\lambda \in (0,1)$.

\section{Experiments}
\subsection{Dataset}
To evaluate the performance of HOSDP, we conduct experiments on the SemEval 2015 Task 18 English datasets \cite{oepen2015semeval}, where all sentences
are annotated with three different formalisms:
DELPH-IN MRS (DM), Predicate-Argument Structure (PAS)  
and Prague Semantic Dependencies (PSD). 
We use the same dataset split as in
previous approaches \cite{almeida2015lisbon,du2015peking}
with 33,964 training sentences from
Sections 00-19 of the Wall Street Journal corpus
\cite{marcus1993building}, 1,692 development sentences
from Section 20, 1,410 sentences from Section 21
as in-domain (ID) test set, and 1,849 sentences sampled
from the Brown Corpus \cite{francis1982frequency} 
as out-of-domain (OOD) test data. For the evaluation,
we use the evaluation script used in \cite{wang2019second}, 
reporting labelled F-measure scores (LF1) (including ROOT arcs)
on the ID and OOD test sets for each formalism as well as the macro-average
over the three of them.

\subsection{Hyperparameters}
The hyperparameter configuration for our final
system is given in Appendix \ref{Hyperparameters}. 
We use 100-dimensional pretrained GloVe embeddings. Only
words or lemmas that occurred 7 times or more
are included in the word and lemma embedding
matrix—including less frequent words appeared to
facilitate overfitting. Character-level word embeddings
are generated using a one-layer unidirectional
LSTM that convolved over three character
embeddings at a time, whose final state is linearly
transformed to be 100-dimensional.

% Following \citet{wang2019second}, we use Adam for optimizing our model, annealing the learning rate by 0.5 for every 10,000 steps, and switched the optimizer to AMSGrad after 5,000 steps without improvement. 
We train the model for at most 100 epochs and terminated training early after 20 epochs with no improvement on the development set.

\begin{table*}[h]  
\centering
\begin{tabular}{lccccccccc}
\toprule   
Models & \multicolumn{2}{c}{DM} &  \multicolumn{2}{c}{PAS} &  \multicolumn{2}{c}{PSD} & \multicolumn{2}{c}{Avg}\\ & ID & OOD & ID & OOD & ID & OOD & ID & OOD\\  

\midrule   
Du et al. \shortcite{du2015peking} & 89.1 & 81.8 &91.3 &87.2 &75.7 &73.3 &85.3 &80.8   \\  
A\&M \shortcite{almeida2015lisbon} & 88.2 & 81.8 & 90.9 & 86.9 & 76.4 & 74.8 & 85.2 & 81.2  \\
PTS17 \shortcite{peng2017deep}: Basic & 89.4 & 84.5 & 92.2 & 88.3 & 77.6 & 75.3 & 86.4 & 82.7\\
PTS17 \shortcite{peng2017deep}: Basic & 90.4 & 85.3 & 92.7 & 89.0 & 78.5 & 76.4 & 87.2 & 83.6\\
WCGL \shortcite{wang2018neural} & 90.3 & 84.9 & 91.7 & 87.6 & 78.6 & 75.9 & 86.9 & 82.8  \\
D\&M \shortcite{dozat2018simpler}: Basic &91.4 & 86.9 & 93.9 & 90.8 & 79.1 & 77.5 & 88.1 & 85.0   \\
MF \shortcite{wang2019second}: Basic & 93.0 & \textbf{88.4} & 94.3 & 91.5 & 80.9 & 78.9 & 89.4 & 86.3   \\
LBP \shortcite{wang2019second}: Basic & 92.9 & \textbf{88.4} & 94.3 & 91.5 & 81.0 & 78.8 & 89.4 & 86.2\\
Lindemann et al. \shortcite{lindemann2019compositional}: Basic & 91.2 & 85.7 & 92.2 & 88.0 & 78.9 & 76.2 & 87.4 & 83.3 \\
Semantic Pointer \shortcite{fernandez2020transition}: Basic & 92.5 & 87.7 & 94.2 & 91.0 & 81.0 & 78.7 & 89.2 & 85.8\\
HOSDP (GCN): Basic & \textbf{93.3} & 88.0 &\textbf{94.8} & 91.1 & \textbf{85.6} &\textbf{83.6}&\textbf{91.2}&\textbf{87.6}\\
HOSDP (GAT): Basic & {93.0} & {87.9} &\textbf{94.8} &\textbf{91.6} & \textbf{85.4} &\textbf{83.3}&\textbf{91.1}&\textbf{87.6}\\
% HOSDP (4th-order): Basic & \textbf{93.9} &\textbf{89.4} &\textbf{95.8} &\textbf{93.4} & \textbf{} &\textbf{}&\textbf{}&\textbf{}\\
\midrule 
D\&M \shortcite{dozat2018simpler}: \scriptsize{+Char+Lemma}& 93.7 & 88.9 & 93.9 & 90.6 & 81.0 & 79.4 & 89.5 & 86.3\\
MF \shortcite{wang2019second}: \scriptsize{+Char+Lemma} & 94.0 & 89.7 & 94.1 & 91.3 & 81.4 & 79.6 & 89.8 & 86.9\\
LBP \shortcite{wang2019second}: \scriptsize{+Char+Lemma} & 93.9 & 89.5 & 94.2 & 91.3 & 81.4 & 79.5 & 89.8 & 86.8\\
Jia et al. \shortcite{jia2020semi}: \scriptsize{+Lemma} & 93.6 & 89.1 & - & - & - & - & - & - \\
Semantic Pointer \shortcite{fernandez2020transition}: \scriptsize{+Char+Lemma} & 93.9 & 89.6 & 94.2 & 91.2 & 81.8 & 79.8 & 90.0 & 86.9\\

HOSDP (GCN): \scriptsize{+Char+Lemma} & \textbf{94.2} & \textbf{90.1} & \textbf{ 94.9} & \textbf{91.4} &\textbf{86.4} & \textbf{84.9} & \textbf{91.8} & \textbf{88.8}\\
HOSDP (GAT): \scriptsize{+Char+Lemma} & \textbf{94.4} & \textbf{89.9} & \textbf{95.0} & \textbf{91.8} &\textbf{86.2} & \textbf{84.6} & \textbf{91.9} & \textbf{88.8}\\
\midrule 
Lindemann et al. \shortcite{lindemann2019compositional}: \scriptsize{+BERT\tiny{large}} & 94.1 & 90.5 &  94.7 & 92.8 & 82.1 & 81.6 & 90.3 & 88.3\\
Lindemann et al. \shortcite{lindemann2020fast}: \scriptsize{+BERT\tiny{large}} & 93.9 & 90.4 & 94.7 & 92.7 &  81.9 & 81.6 & 90.2 & 88.2\\
Semantic Pointer \shortcite{fernandez2020transition}: \scriptsize{+Char+Lemma+BERT\tiny{base}} & 94.4 & 91.0 & 95.1 & 93.4 & 82.6 & 82.0 & 90.7 & 88.8\\
He et al. \shortcite{he2020establishing}: \scriptsize{+Char+Lemma+BERT\tiny{base}+Flair} & 94.6 & 90.8 & \textbf{96.1} & \textbf{94.4} & 86.8 & 79.5 & 92.5 & 88.2\\
HOSDP (GCN): \scriptsize{+Char+Lemma+BERT\tiny{base}}& \textbf{95.1} & \textbf{91.1} & 95.7& 93.2 & \textbf{87.7}&\textbf{87.3}&\textbf{92.8}&\textbf{90.5}\\
HOSDP (GAT): \scriptsize{+Char+Lemma+BERT\tiny{base}} & \textbf{95.3} & \textbf{91.9} & 96.0 & 94.3 & \textbf{87.0}&\textbf{86.7}&\textbf{92.8}&\textbf{91.0}\\
\bottomrule  

\end{tabular}
\caption{Comparison of labeled F1 scores achieved by our model and previous state-of-the-art. The F1 scores of Baseline and our models are averaged over 5 runs.  ID denotes the in-domain (WSJ) test set and OOD denotes the out-of-domain (Brown) test set. +Char, +Lemma, +BERT, and +Flair mean augmenting the token embeddings with character-level, lemma embeddings, BERT embeddings, and Flair embeddings. Semi-SDP only shows the full-supervised result on DM formalism.}
\label{main_result}
\end{table*}

\subsection{Baseline}
We compare HOSDP with previous state-of-the-art
parsers in Table \ref{main_result}. \citet{du2015peking} is a
hybrid model. A\&M is from \citet{almeida2015lisbon}. 
WCGL \cite{wang2018neural} is a neural transition-based model.
PTS17 proposed by \citet{peng2017deep} and
Basic is single task parsing while Freda3 is a multitask
parser across three formalisms.  
D\&M \cite{dozat2018simpler} is a
first-order graph-based model. MF and LBP \cite{wang2019second} 
are a second-order model using mean field variational inference
or loopy belief propagation.
\citet{lindemann2019compositional} and \citet{lindemann2020fast}
are compositional semantic parser for SDP and abstract meaning 
representation. Semantic Pointer \cite{fernandez2020transition}
is a transition-based model using Pointer Network. \citet{jia2020semi} 
is a semi-supervised parser, only the full-supervised result on DM 
formalism is shown in their paper. 
\citet{he2020establishing} uses not only BERT but also contextual string embeddings (called Flair).

\subsection{Main Results}
We group SDP models in three blocks depending on the embeddings provided to the models:
(1) just basic pre-trained GloVe word embeddings and POS tag embeddings (Basic),
(2) character and pre-trained lemma embeddings augmentation (+Char+Lemma) and 
(3) pre-trained BERT embeddings augmentation (+Char+Lemma+BERT).

Table \ref{main_result} presents the comparison of HOSDP
and related studies on the test sets of SemEval 2015 Task 18 English datasets. 
From the results, we have the following observations:
\begin{itemize}
    \item In the basic setting, HOSDP achieves 1.8 and 1.3 averaged LF1 improvements on the ID and OOD test set over the previous best parsers.
     
    \item Adding both the character-level and the lemma embeddings, most models improve performance quite a bit generally.  
    % A reasonable explanation is that many infrequent words     are out of token embedding matrix, character-level and      lemma embeddings make it discriminative.
    HOSDP leads to 1.9 and 1.9 averaged LF1 improvements over the previous best parsers on the ID and OOD test sets.   
    
    \item Adding BERT embedding pushes performance even higher generally. 
    HOSDP outperforms the previous best parsers by 0.3 and 2.2 average LF1 improvements on ID and OOD test sets, respectively. 
    % The reason is that deep contextualized embeddings augment word embeddings.
    
    \item HOSDP makes significant improvements on the PSD formalism, 
    with 4.6 and 4.7 LF1 improvements in the basic setting, 
    4.6 and 5.1 LF1 improvements when character-level and lemma embeddings are added, 0.9 and 5.3 LF1 improvements when BERT embeddings are added on the ID and OOD test sets, respectively.
    
    \item The LF1 scores of all parsers on the PSD is lower than the other two formalisms. Table \ref{statistics} shows part of contrastive statistics of three formalisms. We have noticed that PSD  formalism appears linguistically most fine-grained because it contains the most semantic labels and frames \cite{oepen2015semeval}. This makes PSD  more challenging to predict. However, HOSDP performs better than other first-order and second-order parsers, suggesting that higher-order information is beneficial to SDP.
    
    \item The performances of HOSDP (GCN) and HOSDP (GAT) are almost the same in three embedding settings, demonstrating that both GCNs and GAT are capable to capture higher-order information.
\end{itemize}

In summary, outstanding performances of HOSDP have demonstrated that higher-order information can bring considerable accuracy gains in SDP. In addition, GNNs are capable to capture higher-order information and are effective for higher-order modeling. 

\begin{table}[h]  
\centering
\begin{tabular}{lcccccc}
\toprule   
 & \multicolumn{2}{c}{DM} & \multicolumn{2}{c}{PAS} & \multicolumn{2}{c}{PSD} 
\\ & ID & OOD & ID & OOD & ID & OOD\\  

\midrule   
\# labels & 59 & 47  & 42 & 41 & 91 & 74 \\
\# frames & 297 & 172 & - & - & 5426 & 1208 \\

\bottomrule  
\end{tabular}
\caption{Contrastive statistics of three semantic formalisms.}
\label{statistics}
\end{table}

\begin{figure*}[!ht]
	\centering  %图片全局居中
	\vspace{-0.35cm} %设置与上面正文的距离
% 	\subfigtopskip=2pt %设置子图与上面正文或别的内容的距离
	\subfigbottomskip=2pt %设置第二行子图与第一行子图的距离，即下面的头与上面的脚的距离
	\subfigcapskip=-1pt %设置子图与子标题之间的距离
	\subfigure[DM.ID]{
		\label{DM.ID}
		\includegraphics[width=0.32\textwidth]{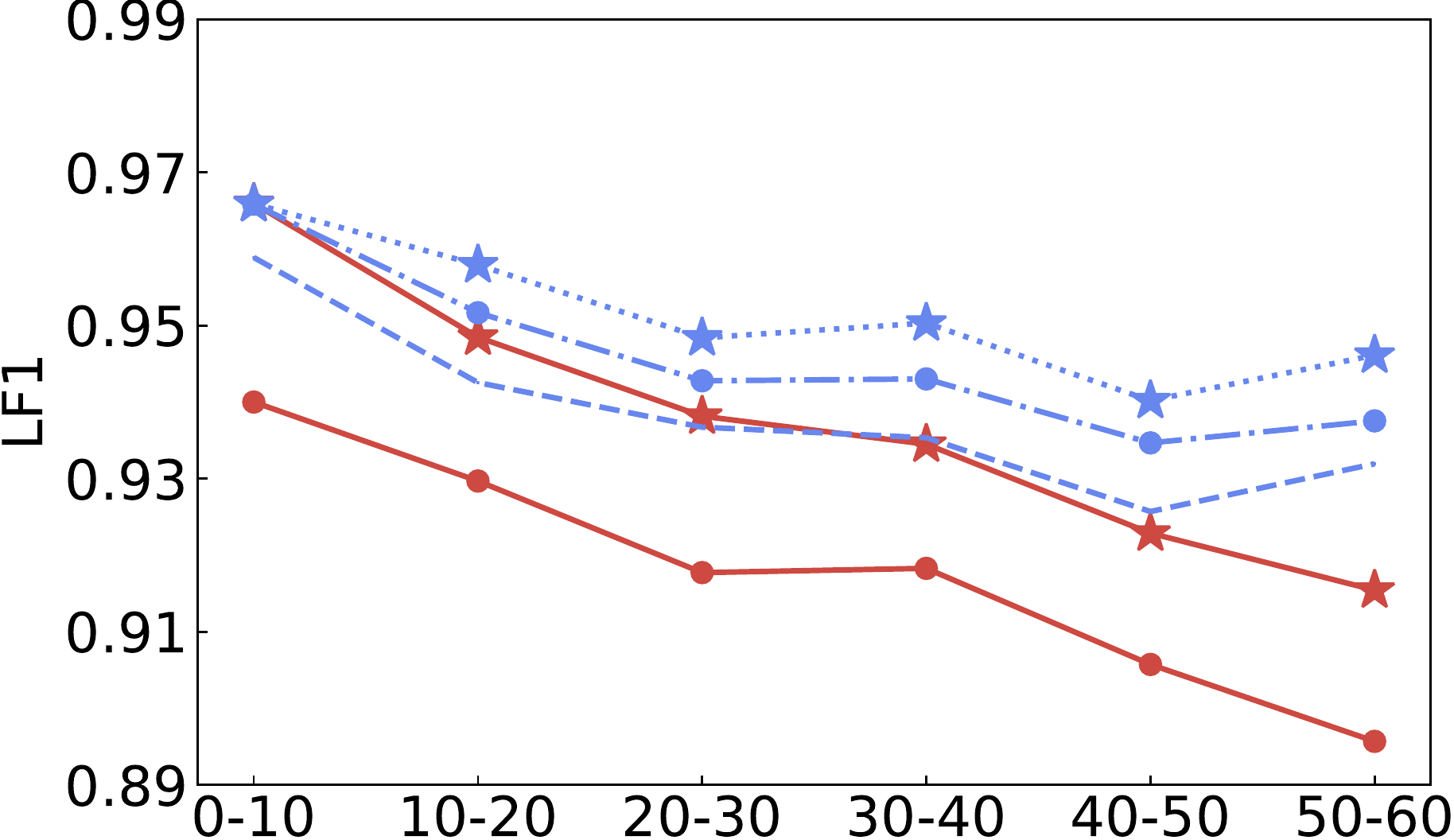}}
	\subfigure[PAS.ID]{
		\label{PAS.ID}
		\includegraphics[width=0.32\textwidth]{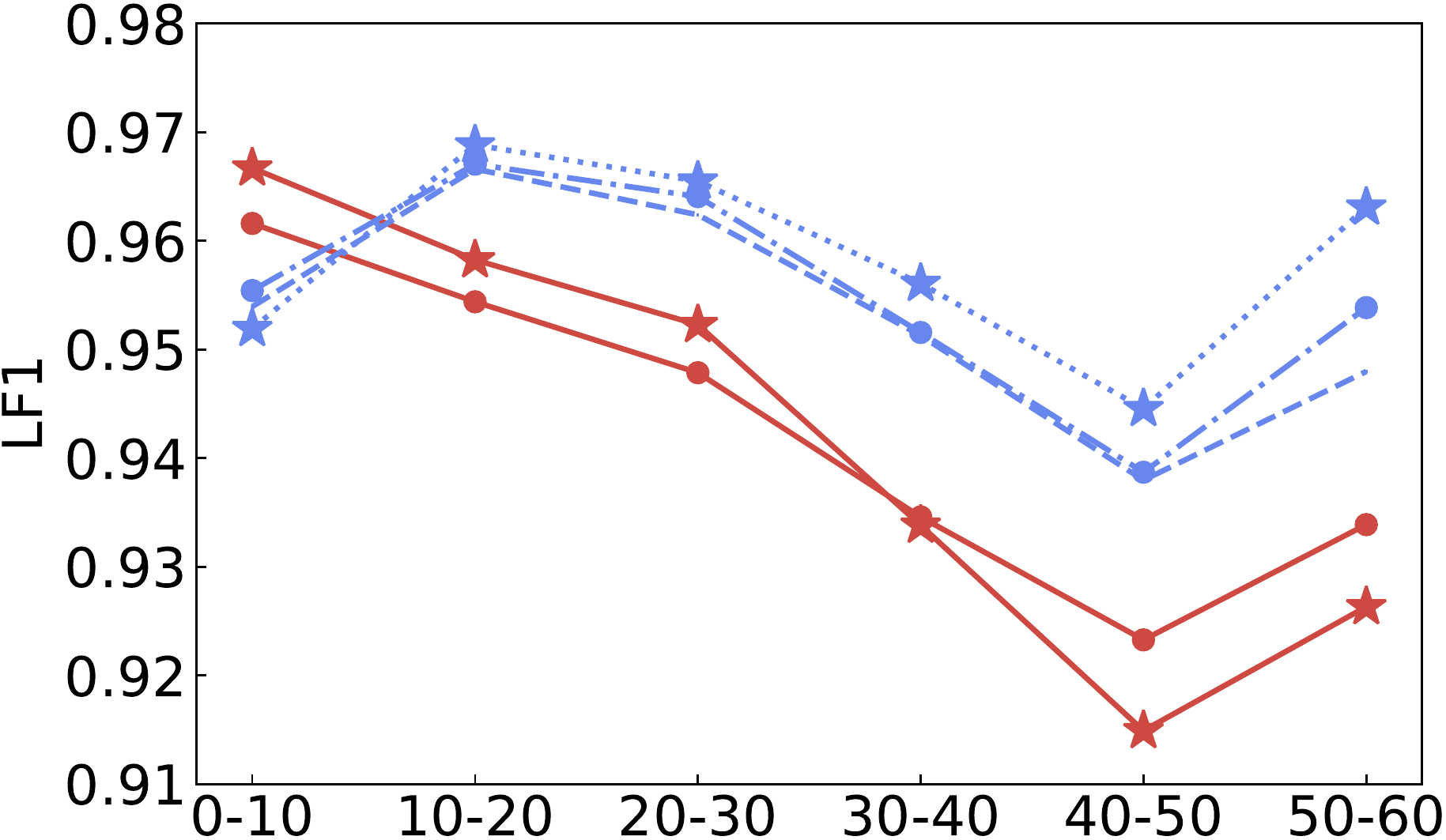}}
	\subfigure[PSD.ID]{
		\label{PSD.ID}
		\includegraphics[width=0.32\textwidth]{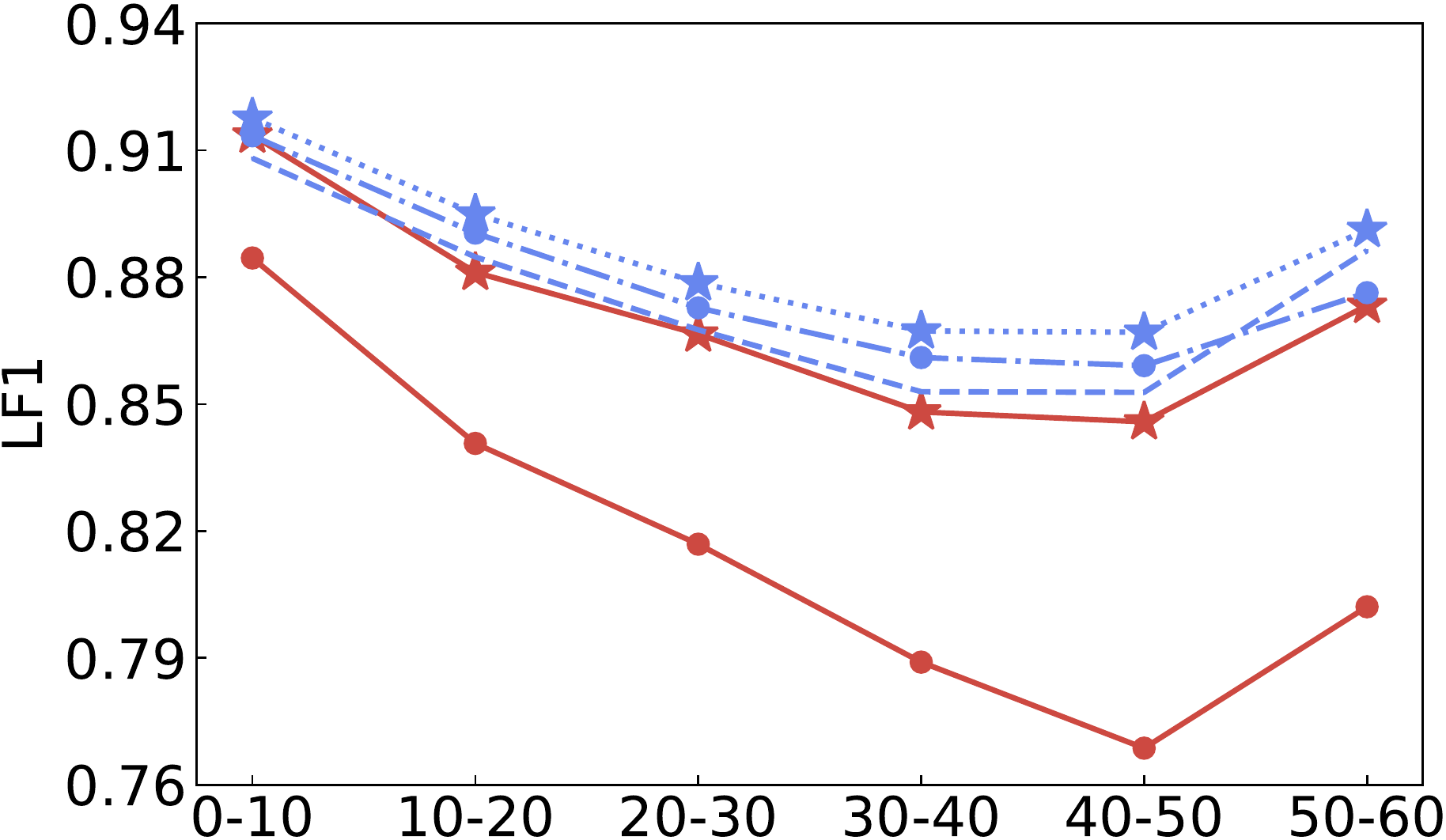}}
	\subfigure[DM.OOD]{
		\label{DM.OOD}
		\includegraphics[width=0.32\textwidth]{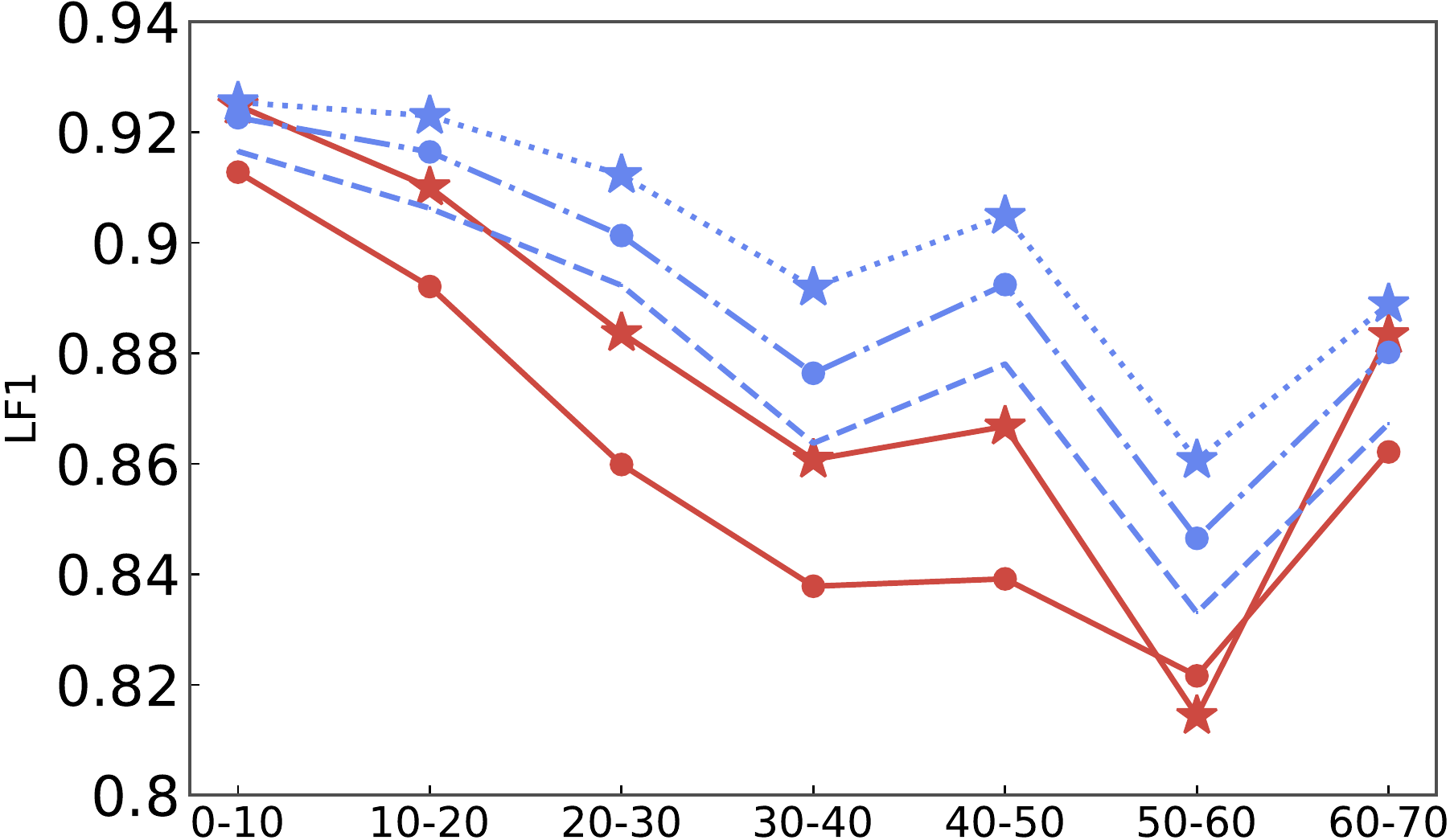}}
	\subfigure[PAS.OOD]{
		\label{PAS.OOD}
		\includegraphics[width=0.32\textwidth]{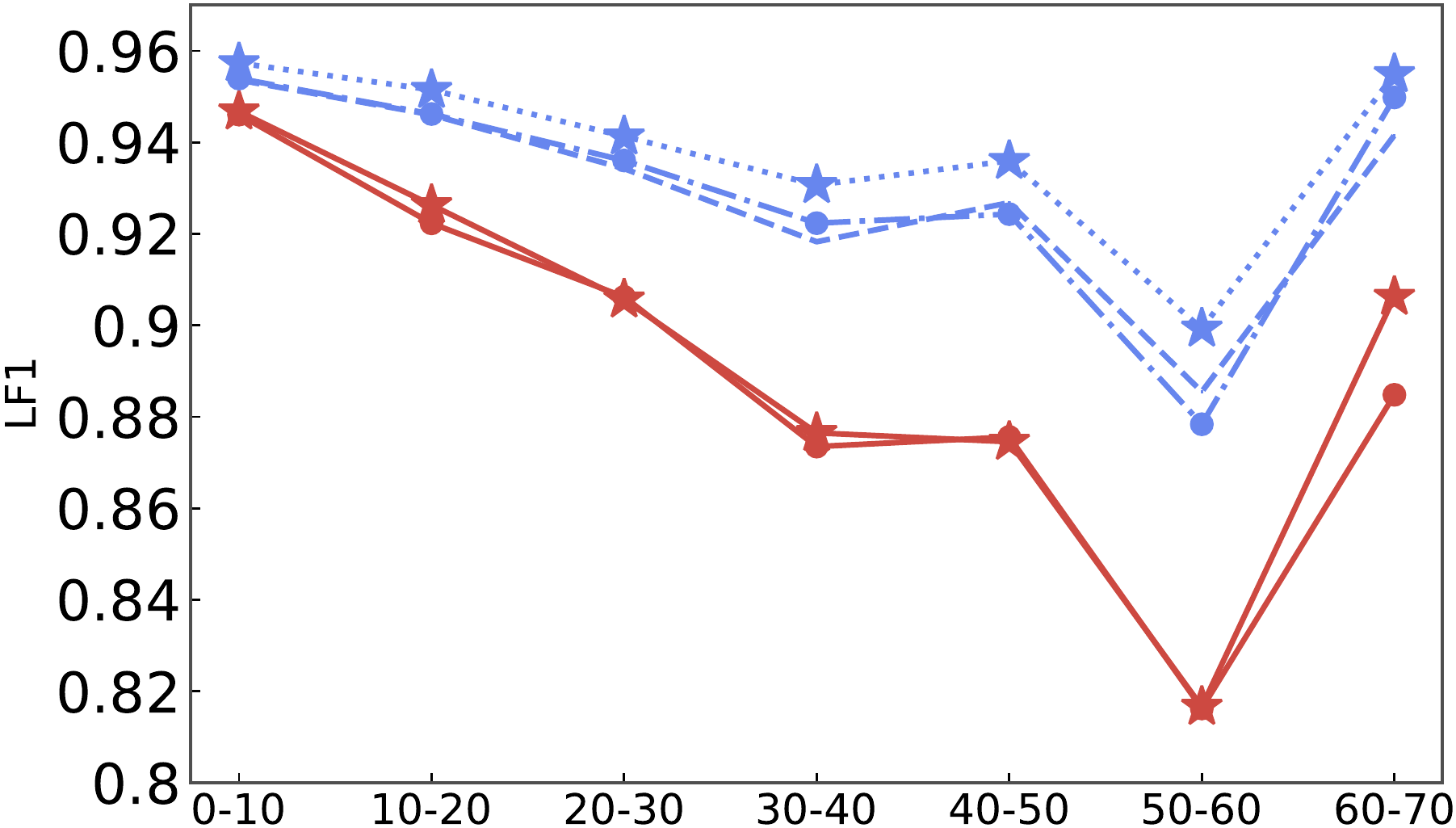}}	
	\subfigure[PSD.OOD]{
		\label{PSD.OOD}
		\includegraphics[width=0.32\textwidth]{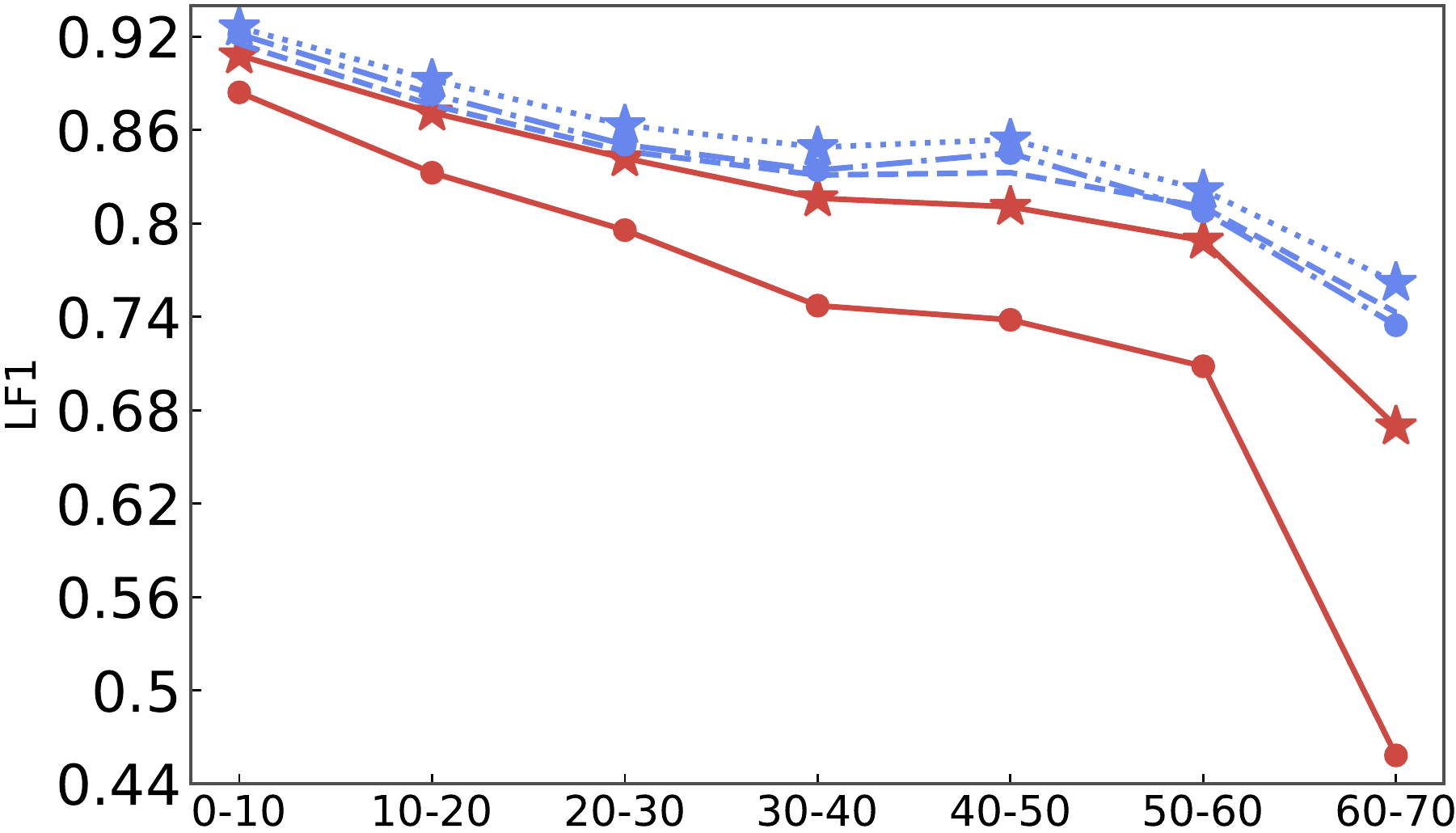}}
	\subfigure{
		\label{plot_lagend}
		\includegraphics[width=0.4\textwidth]{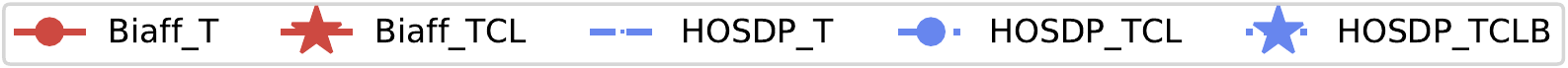}}	
	\caption{LF1 of different sentence lengths in 3 formalisms. 
	$\star$-T represents that only the POS tag is used as feature.
	$\star$-TCL represents that the POS tag, character-level and 
	lemma embeddings are used as features. $\star$-TCLB represents that 
	the POS tag, character-level, lemma and BERT embeddings 
	are used as features.}
	\label{sent_len_fig}
\end{figure*}

\subsection{Analysis}

% \subsubsection*{Error analysis}

\subsubsection*{Performance on Different Sentence Length}
We want to study the impact of sentence lengths. 
The ID and OOD test sets of the three formalisms 
are split with 10 tokens range. The ID test set has 6 groups and OOD
has 7. HOSDP (GCN) and biaffine parser are evaluated on them.
The results on diffenent groups are shown in Figure \ref{sent_len_fig}. 

The results show that HOSDP outperforms the biaffine parser on different groups with the same embedding configuration, except for slightly lower on the first group (0-10 tokens) on ID test set of PAS formalism.
Furthermore, HOSDP that only utilizes POS tag embedding outperforms biaffine parse that uses POS tag, character-level, and lemma embeddings, when sentences get longer, especially when sentences are longer than 30. It turns out that higher-order information is favorable for longer sentences since higher-order dependency relationships are more prevalent in longer sentences.

\subsubsection*{Case study}
\begin{figure}[!ht]
	\centering  %图片全局居中
	\vspace{-0.35cm} %设置与上面正文的距离
	\subfigtopskip=2pt %设置子图与上面正文或别的内容的距离
	\subfigbottomskip=2pt %设置第二行子图与第一行子图的距离，即下面的头与上面的脚的距离
	\subfigcapskip=-2pt %设置子图与子标题之间的距离
	\subfigure[Parsing result of biaffine parser]{
		\label{biaff_result}
		\includegraphics[width=0.48\textwidth]{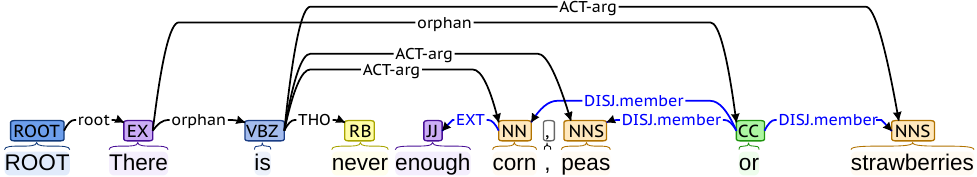}}
	\subfigure[Parsing result of HOSDP (GCN)]{
		\label{HOSDP_result}
		\includegraphics[width=0.48\textwidth]{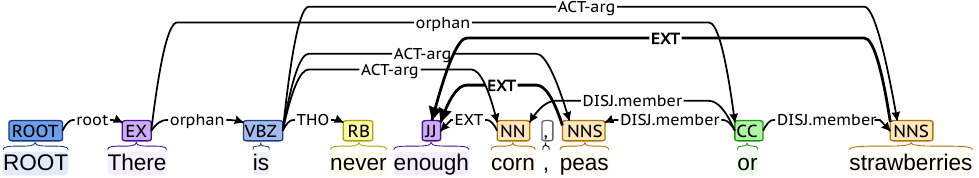}}
	\caption{Parsing results of biaffine parser and HOSDP, 
    for the sentence (sent\_id=40504035 in OOD of PSD formalism). Some irrelevant dependency edges are hidden. }
	\label{sent_parse_fig}
\end{figure}

We provide a parsing example to show why HOSDP benefits from higher-order information. Figure \ref{sent_parse_fig} shows the parsing results of biaffine parser (Figure \ref{biaff_result}) and HOSDP (GCN) (Figure \ref{HOSDP_result}), for sentence (sent\_id=40504035, in OOD of PSD formalism): \emph{There is never enough corn, peas or strawberries}. Both of two parsers are trained in the basic embedding setting.

In the gold annotation, three words \emph{corn}, \emph{peas} and \emph{strawberries} are three members of the disjunctive word \emph{or}. In addition, the word \emph{enough} has three dependency edges labeled \emph{EXT} with them.
In the result of biaffine parser, only dependency edge between \emph{enough} and \emph{corn} is identified, the remaining two are not.
In HOSDP (GCN), given the initial SDG predicted by biaffine parser, words \emph{corn}, \emph{peas} and \emph{strawberries} aggregate the higher-order information of \emph{or} and \emph{is} ($1^{st}$-order), \emph{there} ($2^{nd}$-order) and \emph{ROOT} ($3^{rd}$-order). The word \emph{enough} aggregate the higher-order information of the \emph{corn} ($1^{st}$-order), \emph{is} and \emph{or} ($2^{nd}$-order), \emph{there} ($3^{rd}$-order), and \emph{ROOT} ($4^{th}$-order). Dependent-word \emph{enough} and three head-words \emph{corn}, \emph{peas} and \emph{strawberries} aggregate information of four common words (\emph{ROOT}, \emph{There}, \emph{is} and \emph{or}). Therefore the representations of them with higher-order information bring global evidence into decoders’ final decision.
As a result, it is effortless for HOSDP to identify that there are also two dependency edges labeled \emph{EXT} between dependent-word \emph{enough} and head-words \emph{peas} and \emph{strawberries}.

\section{Conclusions}
We propose a higher-order semantic dependency parser, which employs the GNNs to capture higher-order information. Experimental results show that HOSDP outperforms previous best one on the SemEval 2015 Task 18 English datasets. In addition, HOSDP shows more advantage in longer sentence and complex semantic formalism. In the future, we would like to apply graph structure learning models to learn adjacency matrix, rather than depending on the results of vanilla parser. 

% \section*{Acknowledgments}
% This work was sponsored by the Foundation of Jiangsu Provincial Double-Innovation Doctor Program (Grant No. JSSCBS20210507) and 
% NUPTSF (Grant No. NY220176).

\clearpage
\bibliographystyle{named}
\footnotesize
\bibliography{ijcai22}

\clearpage
\appendix
\section{Hyperparameter values}\label{Hyperparameters}
  \begin{table}[!hb]
	\centering
	\begin{tabular}{ccc}
		\toprule
		 Layer & Hyper-parameter & Value\\
		 \midrule
		 \multirow{1}*{Input} & Glove/POS/Lemma/Char/BERT & 100  \\
		\midrule
		\multirow{3}*{LSTM} & layers & 3  \\
		~ & hidden size & 400  \\
		~ & dropout & 0.33  \\
		\midrule
		\multirow{4}*{GNN} & GCN layers & 3  \\
		~ & GAT heads & 8  \\
		~ & GAT $\alpha$ & 0.2  \\
		~ & GCN/GAT dropout & 0.33  \\
		\midrule
		\multirow{2}*{MLP} & edge-head/label-head hidden size & 600\\
		~ & edge-dep/label-dep hidden size & 600 \\
		\midrule
		\multirow{5}*{Trainer} & optimizer  & Adam\\
		~ & learning rate & $1e^{-2}$ \\
		~ & Adam ($\beta_1$, $\beta_2$) & (0.95, 0.95) \\
		~ & decay rate & 0.75 \\
		~ & decay step length & 5000  \\
		~ & $\lambda$ & 0.1  \\
		\bottomrule
	\end{tabular}
	\caption{Final hyperparameter configuration.}
    \label{tab_parameters}
\end{table}

\end{document}